\documentclass[10pt,twocolumn,letterpaper]{article}

\usepackage{cvpr}              % To produce the CAMERA-READY version
\usepackage{xspace}
\newcommand{\ours}{ST-SfM\xspace}

\usepackage{multirow}
\usepackage{tikz}
\usetikzlibrary{positioning,arrows.meta,fit,backgrounds}

\definecolor{cvprblue}{rgb}{0.21,0.49,0.74}
\usepackage[pagebackref,breaklinks,colorlinks,allcolors=cvprblue]{hyperref}

\def\confName{CVPR}
\def\confYear{2027}

\title{ST-SfM: Spatio-Temporal Structure-from-Motion\\for Initialising 4D Gaussian Splatting}

\author{
    Harshit\\
    IT University of Copenhagen\\
    Copenhagen, Denmark\\
    \texttt{\small hano@itu.dk}
    \and
    Sami S. Brandt\\
    IT University of Copenhagen\\
    Copenhagen, Denmark\\
    \texttt{\small sambr@itu.dk}
}

\begin{document}
\maketitle
\begin{abstract}
   We propose ST-SfM, a novel pipeline for efficient and accurate initialization of 4D Gaussian Splatting (4DGS) using sparse, geometry-consistent spatio-temporal features derived from multi-view video. Unlike current 4DGS methods that rely on structure-from-motion (SfM) cloud or random initialization, our approach leverages structured visual information to guide the initial distribution of Gaussians in space-time. We treat each input video stream as a (x, y, t) volume and detect salient static and dynamic interest points. Static structure is detected once per camera on a temporal-median image, filtered for temporal stability, and matched along epipolar lines; under 2\% of the kept static keypoints fall on moving objects. Motion is detected as space-time interest points (STIPs) in the video volume, described with HOG3D, and matched only between candidates that agree both epipolarly and temporally ($|\Delta t|\le 1$ frame). A conflict-aware union-find assembles tracks while rejecting inconsistent matches, and RANSAC triangulation over all view pairs yields static 3D points and dynamic 4D points $(x,y,z,t)$ with sub-frame timestamps. On N3DV scenes, \ours reconstructs 39k-113k dynamic points per scene at 0.65-0.83\,px mean reprojection error; detection takes about 6\,s per camera for a 300-frame video volume, and matching and triangulation take 2-20\,s per scene. Our method provides an efficient alternative to heuristic initialization strategies, opening a new direction for structured, spatio-temporal feature-guided dynamic scene reconstruction.
\end{abstract}

\section{Introduction}
\label{sec:intro}

3D Gaussian splatting (3DGS)~\cite{kerbl3Dgaussians} and its dynamic extensions~\cite{yang2024real,wu20244d,luiten2024dynamic,li2024spacetime} have made photorealistic, real-time novel-view synthesis of dynamic scenes practical.
All of them refine a set of primitives by gradient descent and grow the set by heuristic densification, so the primitives they start from matter.
For static scenes the answer is settled: initialise from the sparse SfM cloud that already came with the camera poses~\cite{schoenberger2016sfm}.
For dynamic scenes there is no such standard.
Implementations use either points sampled at random in the camera frusta, or an SfM (sometimes multi-view stereo) cloud of the \emph{first frame}, with a broad, uniform distribution over time~\cite{yang2024real,wu20244d}.
A first-frame cloud describes one instant. It places static and moving structure at their frame-0 positions and has nothing to say about anything that happens later in the sequence.

The information that is missing, \emph{where and when something moves}, has been studied for twenty years in action recognition.
Space-time interest points (STIPs)~\cite{laptev2005space,dollar2005behavior,willems2008efficient} are extrema of a spatio-temporal scale space in the $(x,y,t)$ video volume, and are well localised in both space and time.
Spatio-temporal descriptors~\cite{scovanner20073,klaser2008spatio} make them matchable.
For a synchronised, calibrated multi-camera rig, a STIP seen in two cameras at the same instant is a correspondence like any other, and triangulating it gives a 3D point \emph{with a time stamp}.
STIPs fire only where the image changes, however, so they say almost nothing about the static scene.

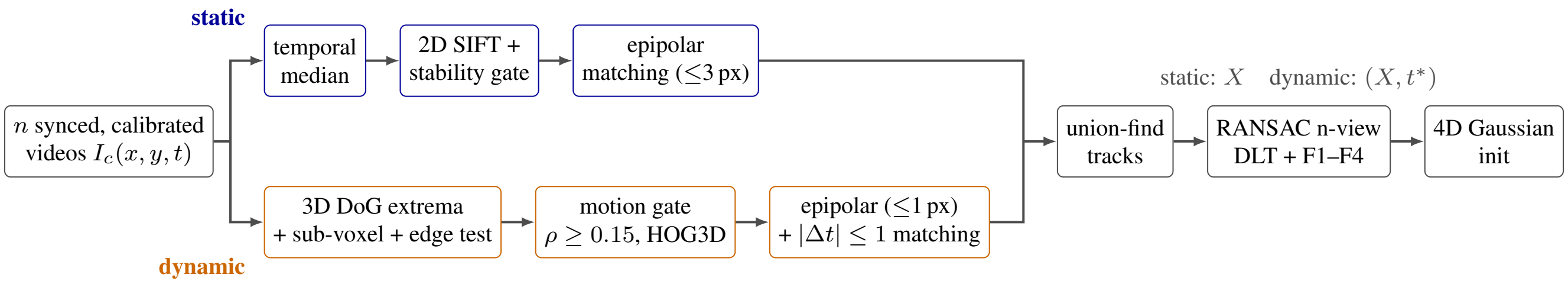
\begin{figure*}[t]
  \centering
  \resizebox{\linewidth}{!}{\begin{tikzpicture}[
      font=\footnotesize,
      box/.style={draw, rounded corners=2pt, align=center, minimum height=8.5mm, inner sep=3pt, fill=white},
      stat/.style={box, draw=blue!60!black},
      dyn/.style={box, draw=orange!80!black},
      both/.style={box, draw=black!70},
      arr/.style={-{Latex[length=1.6mm]}, thick, black!70},
      node distance=4mm and 4mm]
    \node[both] (vid) {$n$ synced, calibrated\\videos $I_c(x,y,t)$};
    % static row
    \node[stat, above right=1mm and 6mm of vid] (med) {temporal\\median};
    \node[stat, right=of med] (sift) {2D SIFT +\\stability gate};
    \node[stat, right=of sift] (smatch) {epipolar\\matching ($\le$3\,px)};
    % dynamic row
    \node[dyn, below right=1mm and 6mm of vid] (dog) {3D DoG extrema\\+ sub-voxel + edge test};
    \node[dyn, right=of dog] (gate) {motion gate\\ $\rho\ge0.15$, HOG3D};
    \node[dyn, right=of gate] (dmatch) {epipolar ($\le$1\,px)\\+ $|\Delta t|\le1$ matching};
    % shared
    \coordinate (j) at ([xshift=4mm]dmatch.east |- vid);
    \node[both, anchor=west] (tracks) at ([xshift=4mm]j) {union-find\\tracks};
    \node[both, right=of tracks] (tri) {RANSAC n-view\\DLT + F1--F4};
    \node[both, right=of tri] (init) {4D Gaussian\\init};
    \draw[arr] (vid.east) -- ++(2mm,0) |- (med.west);
    \draw[arr] (vid.east) -- ++(2mm,0) |- (dog.west);
    \draw[arr] (med) -- (sift); \draw[arr] (sift) -- (smatch);
    \draw[arr] (dog) -- (gate); \draw[arr] (gate) -- (dmatch);
    \draw[thick, black!70] (smatch.east) -| (j);
    \draw[thick, black!70] (dmatch.east) -| (j);
    \draw[arr] (j) -- (tracks.west);
    \draw[arr] (tracks) -- (tri); \draw[arr] (tri) -- (init);
    \node[above=0.5mm of tri, align=center, text=black!70] {static: $X$\quad dynamic: $(X,t^\ast)$};
    \node[left=1mm of med, text=blue!60!black, anchor=south east, yshift=3mm] {\textbf{static}};
    \node[left=1mm of dog, text=orange!80!black, anchor=north east, yshift=-3mm] {\textbf{dynamic}};
  \end{tikzpicture}}
  \caption{\textbf{\ours overview.} Each camera's video is processed by two branches. The static branch (blue) runs once per camera on a temporal-median image. The dynamic branch (orange) runs on the full $(x,y,t)$ volume. Both branches share calibrated guided matching, consistent multi-view tracks and robust n-view triangulation. Static points become time-invariant Gaussians; dynamic points become Gaussians centred at their fused time $t^\ast$.}
  \label{fig:pipeline}
\end{figure*}

We therefore split the problem into two branches (\cref{fig:pipeline}).
The \emph{static} branch detects 2D SIFT~\cite{lowe2004distinctive} keypoints once per camera on a temporal-median image and rejects any whose patch changes over time.
The \emph{dynamic} branch detects STIPs with a difference-of-Gaussians detector on the video volume and keeps only those whose gradient energy is mainly temporal.
Both branches use the known calibration more strongly than generic SfM matching does.
A match is searched only among candidates within 1--3\,px of the epipolar line in both images and, for STIPs, within one frame in time.
Matches are chained into multi-view tracks by a union-find that refuses inconsistent merges, and each track is triangulated by RANSAC over all of its view pairs.
The result is a static cloud and a dynamic cloud of $(x,y,z,t)$ points, which we convert directly into 4D Gaussians with time centres, temporal extents and a static flag.

\noindent Our contributions are:
\begin{itemize}
  \item \textbf{\ours}, a sparse spatio-temporal SfM for synchronised multi-camera video that reconstructs a sparse 4D structure cloud, with sub-pixel accuracy.
  \item \textbf{Epipolar- and time-guided STIP matching with consistent n-view tracks}, which replaces pairwise triangulation and removes redundant points.
  \item \textbf{A 4D Gaussian initialisation from \ours}, evaluated on three N3DV scenes.
\end{itemize}

\section{Related Work}
\label{sec:related}

\paragraph{Dynamic radiance fields and Gaussians.}
Dynamic NeRFs model time with deformation fields or factorised space-time grids~\cite{li2022neural,fridovich2023k,cao2023hexplane}.
Gaussian methods either deform a canonical set of 3D Gaussians~\cite{wu20244d}, track persistent Gaussians frame by frame~\cite{luiten2024dynamic}, attach temporal opacity and polynomial motion to each Gaussian~\cite{li2024spacetime}, or treat each primitive as a 4D Gaussian in space-time whose conditional is a 3D Gaussian at each instant~\cite{yang2024real}.
We build on the last of these, 4DGS~\cite{yang2024real}, because a 4D Gaussian has an explicit time centre and temporal extent.
A time-stamped point therefore maps directly onto it.
All of these methods start from either random points or a first-frame SfM/MVS cloud. Our contribution is to the starting point, not to the representation.

\paragraph{Initialisation of Gaussian splatting.}
For static scenes, 3DGS is sensitive to its SfM initialisation: random initialisation loses quality, especially on unbounded backgrounds~\cite{kerbl3Dgaussians}.
This has motivated strategies that relax the dependence on accurate points~\cite{jung2024relaxing} or replace SfM with learned dense geometry~\cite{fan2024instantsplat}.
For dynamic scenes the question has received less attention. Standard practice is a frame-0 COLMAP cloud for multi-view video~\cite{li2022neural,yang2024real}.
We give the dynamic part of the scene its own spatio-temporal initialisation and keep a static branch for the rest.

\paragraph{Structure from motion.}
Incremental~\cite{snavely2006photo,schoenberger2016sfm} and global~\cite{pan2024glomap} SfM recover poses and sparse structure from 2D features~\cite{lowe2004distinctive}, with robust estimation~\cite{fischler1981random,torr2000mlesac} and multi-view triangulation~\cite{hartley2003multiple}.
They assume a static scene, and moving points are treated as outliers.
We assume the opposite regime: poses and intrinsics are known (as in all multi-view video benchmarks), and the goal is the structure, including the moving part.
Reconstructing moving points from non-simultaneous views is classical trajectory triangulation~\cite{avidan2000trajectory,park20103d}. Sparse dynamic reconstruction has also been addressed for unsynchronised cameras~\cite{zheng2015sparse}.
With synchronised cameras, the dynamic problem reduces to triangulating correspondences that are simultaneous, provided one can find them.

\paragraph{Space-time interest points.}
Laptev~\cite{laptev2005space} extended the Harris detector to the video volume. Dollár \etal~\cite{dollar2005behavior} used separable temporal Gabor filters for denser responses, and Willems \etal~\cite{willems2008efficient} a scale-invariant Hessian detector.
3D SIFT~\cite{scovanner20073} and HOG3D~\cite{klaser2008spatio} describe the surrounding cuboid with histograms of 3D gradient orientations.
These tools were developed to recognise actions from a single view.
To our knowledge they have not been used as cross-view correspondences for metric 4D reconstruction, which is how we use them.

\section{Method}
\label{sec:method}

\paragraph{Setting.}
We are given $n$ synchronised videos $I_c(x,y,t)$, $c=1..n$, $t=0..T{-}1$, with known intrinsics $K_c$ and poses $(R_c, \mathbf{t}_c)$, \ie projection matrices $P_c = K_c[R_c\,|\,\mathbf{t}_c]$ and pairwise fundamental matrices $F_{ab}$.
This is the setting of multi-view video benchmarks such as Neural 3D Video (N3DV)~\cite{li2022neural}, whose poses come from COLMAP.
We output a static point set $\{X_i\}$ and a dynamic set $\{(X_j, t^\ast_j)\}$, each point with its multi-view observations, and convert both into a 4D Gaussian initialisation.

\subsection{Dynamic branch: space-time interest points}
\label{sec:dynamic}

\paragraph{Detection.}
Each video is converted to a grey-level volume $V_c\in\mathbb{R}^{T\times H\times W}$ and smoothed with separable 3D Gaussians $G_\sigma$ at scales $\sigma\in\{0.5,1,1.6,2.4\}$, isotropic in space and time.
Candidates are local extrema of the difference-of-Gaussians stack
\begin{equation}
  D_k = G_{\sigma_{k+1}} \ast V_c - G_{\sigma_k} \ast V_c
\end{equation}
over the $3\times3\times3$ spatio-temporal neighbourhood and the adjacent scales, with $|D_k|\ge 0.05$.
The extremum test is a \texttt{max\_pool3d} comparison on the GPU, and the volume is processed in temporal chunks with a halo.
Following Lowe~\cite{lowe2004distinctive}, we refine every extremum by fitting a quadratic to $D$, giving sub-pixel $(x,y)$ and \emph{sub-frame} $t$. We also reject edge-like responses with the spatial Hessian ratio test ($r=10$).

\paragraph{Motion gate.}
A DoG extremum in $(x,y,t)$ is not necessarily motion: static corners under flicker or noise also respond.
We keep a candidate only if the temporal share of its gradient energy over a $5^3$ cuboid $\mathcal{N}$ is large enough:
\begin{equation}
  \rho = \frac{\sum_{\mathcal{N}} I_t^2}{\sum_{\mathcal{N}} \left(I_x^2 + I_y^2 + I_t^2\right)} \;\ge\; 0.15 .
  \label{eq:rho}
\end{equation}

\paragraph{Description.}
Each STIP is described with HOG3D~\cite{klaser2008spatio}: a $3\times3\times3$ grid of $4^3$-voxel cells, with 3D gradient orientations quantised onto the 20 faces of an icosahedron, giving 540 dimensions.
In a preliminary descriptor study, HOG3D separated true from false correspondences better than 3D SIFT variants~\cite{scovanner20073}, at about a quarter of the cost.

\subsection{Static branch}
\label{sec:static}

A static surface has almost no temporal structure, so most static texture never becomes an $(x,y,t)$ extremum.
We detect it in 2D instead.
For each camera we compute the per-pixel temporal median $M_c$ over all frames, which votes out moving foreground, and run SIFT~\cite{lowe2004distinctive} on $M_c$.
A keypoint can still lie on something that is static only most of the time, such as a person standing still.
We compare its $9\times9$ zero-mean patch in every frame with the zero-mean median patch, and count the frame as \emph{changed} if they differ by more than 12 grey levels.
The keypoint is kept only if at most 5\% of frames are changed.
Removing the mean makes the test insensitive to global exposure drift.
As a proxy for points on movers, we count kept static keypoints with at least 50 STIPs within 15\,px. Across the five N3DV scenes this proxy is 0.15--1.51\% (\cref{tab:recon}).

\subsection{Calibrated, time-guided matching}
\label{sec:matching}

Generic SfM matching compares descriptors first and checks geometry afterwards.
With known calibration we reverse the order.
For a camera pair $(a,b)$ and keypoints $\mathbf{x}_a, \mathbf{x}_b$, let
\begin{equation}
  d_{\mathrm{epi}} = \max\!\big(d(\mathbf{x}_b, F_{ab}\tilde{\mathbf{x}}_a),\; d(\mathbf{x}_a, F_{ab}^\top\tilde{\mathbf{x}}_b)\big)
\end{equation}
be the symmetric point-to-epipolar-line distance in pixels.
A pair is a \emph{candidate} if $d_{\mathrm{epi}}\le\epsilon$, and, for STIPs, if $|t_a - t_b|\le\delta$ in sub-frame time.
We use $\epsilon=3$\,px for static points and $\epsilon=1$\,px for STIPs, and $\delta=1$ frame.
Lowe's ratio test (0.8 static, 0.9 dynamic), a maximum descriptor distance and a mutual nearest-neighbour check are then applied \emph{among candidates only}.
Because the epipolar and temporal gates remove almost every wrong candidate before descriptors are compared, the ratio test has little left to do. The thresholds are robust as a result (\cref{sec:sweep}).
All camera pairs are matched on the GPU; the static branch is matched once, not once per frame.

\subsection{Consistent multi-view tracks}
\label{sec:tracks}

Pairwise triangulation, as in our earlier two-view prototype, turns a point seen by $k$ cameras into up to $k(k{-}1)/2$ slightly different points.
We instead build tracks with a union-find over (camera, keypoint) nodes.
Pairwise matches are merged greedily in order of descriptor distance.
A merge is \emph{refused} if the resulting track would contain two keypoints from the same camera, which no single 3D point can explain.
For dynamic tracks, a merge is also refused if the track would span more than 1.5 frames, which stops chains of $|\Delta t|\le\delta$ matches from drifting in time.

\subsection{Robust n-view triangulation}
\label{sec:triangulation}

Each track is triangulated in four steps, vectorised over all tracks with the same length.
\emph{(i) Hypotheses:} a two-view DLT~\cite{hartley2003multiple} in normalised camera coordinates for every pair of the track's observations, each scored by MSAC~\cite{torr2000mlesac} on pixel reprojection error over all of the track's views (inlier threshold 3\,px).
\emph{(ii) Refit:} an n-view DLT on the best hypothesis' inliers.
\emph{(iii) Refine:} Gauss--Newton on pixel reprojection error, followed by re-classification of inliers.
\emph{(iv) Filters}, generalising the four filters of the two-view prototype to $n$ views:
F1 the point is finite;
F2 its depth is positive and inside every inlier camera's scene bounds (with a 50\% margin);
F3 the maximum pairwise ray angle among inlier views is at least $2^\circ$;
F4 the RMSE over inlier views is at most 2\,px.
Each dynamic point receives the fused time $t^\ast$, the mean sub-frame time of its inlier observations.

\paragraph{Synchronisation check.}
A camera that is out of sync produces dynamic observations that are consistently rejected as outliers by (i).
For every camera we record the fraction of its dynamic observations that are outliers, and flag cameras far above the median.

\subsection{From \ours to 4D Gaussians}
\label{sec:init}

We initialise 4DGS~\cite{yang2024real}, in which each primitive is a 4D Gaussian over $(x,y,z,t)$ and its marginal over $t$ acts as a temporal opacity.
\ours points are mapped into the 4DGS training frame by a similarity fitted to the two sets of camera centres. We verify that its residual is numerically zero (rotation residual $<0.1^\circ$).
\emph{Static points} get a time centre at mid-sequence and a temporal standard deviation of twice the sequence duration, so that their temporal marginal stays above 0.97 over the whole sequence.
Their spatial scale is the 3-nearest-neighbour distance among static points, as in 3DGS~\cite{kerbl3Dgaussians}, and their colour is the median over five training frames.
\emph{Dynamic points} get time centre $t^\ast$ and temporal standard deviation $\max(\sigma_t, \Delta t_{\mathrm{inl}}, 5\text{ frames})/\mathrm{fps}$, where $\sigma_t$ is the detector scale and $\Delta t_{\mathrm{inl}}$ the time span of the inlier observations (median 0.167\,s on \texttt{coffee\_martini}).
Their spatial scale is the 3-nearest-neighbour distance among dynamic points within $\pm3$ frames of $t^\ast$. A neighbour search over all times would find the same moving object at other instants and produce needle-thin Gaussians.
Their colour is sampled from the training image of their first inlier observation.
Everything else follows the 4DGS defaults, including densification and the periodic opacity reset, which we found to be essential (\cref{sec:ablations}).

\section{Experiments}
\label{sec:experiments}

\paragraph{Data.}
We use the five N3DV~\cite{li2022neural} scenes \texttt{coffee\_martini}, \texttt{cook\_spinach}, \texttt{cut\_roasted\_beef}, \texttt{flame\_steak} and \texttt{sear\_steak}.
Each has 18--21 synchronised cameras, 300 frames at 30\,fps and 2704$\times$2028 pixels, which we process at half resolution (1352$\times$1014, as in 4DGS training). Camera poses are decoded from the LLFF \texttt{poses\_bounds.npy} files.
We found that a common decoding treats the stored axes [down, right, back] as OpenGL [right, up, back], which rolls every camera by $90^\circ$. With the corrected decoding, the cameras agree with the 4DGS training cameras to $<0.1^\circ$.

\subsection{Reconstruction}
\label{sec:recon}

\begin{table*}[t]
  \centering
  \small
  \setlength{\tabcolsep}{4.5pt}
  \begin{tabular}{@{}l c rrrr rr rr c@{}}
    \toprule
     & & \multicolumn{4}{c}{Detection (per camera)} & \multicolumn{2}{c}{Static points} & \multicolumn{2}{c}{Dynamic points} & \\
    \cmidrule(lr){3-6}\cmidrule(lr){7-8}\cmidrule(lr){9-10}
    Scene & Cams & STIPs & Static & On-mover & Time & \# & px (mean / p95) & \# & px (mean / p95) & Views/pt \\
    \midrule
    \texttt{coffee\_martini} & 18 & 30.0k & 2,519 & 1.33\% & 5.4\,s & 5,022 & 0.59 / 1.48 & 93,316 & 0.65 / 1.73 & 3.9 \\
    \texttt{cook\_spinach}       & 21 & 34.4k & 1,059 & 0.93\% & 6.2\,s & 3,329 & 0.58 / 1.56 & 113,411 & 0.78 / 1.94 & 4.1 \\
    \texttt{cut\_roasted\_beef}  & 20 & 22.8k &   952 & 1.51\% & 5.5\,s & 2,858 & 0.59 / 1.56 & 76,492 & 0.71 / 1.84 & 3.9 \\
    \texttt{flame\_steak}        & 21 & 27.0k &   981 & 0.15\% & 5.4\,s & 3,100 & 0.57 / 1.56 & 85,919 & 0.83 / 2.02 & 4.7 \\
    \texttt{sear\_steak}         & 21 & 11.1k & 1,051 & 0.21\% & 5.4\,s & 3,320 & 0.57 / 1.55 & 38,502 & 0.77 / 1.90 & 4.4 \\
    \bottomrule
  \end{tabular}
  \caption{\textbf{\ours reconstruction} with all cameras.
  ``On-mover'' is the fraction of kept static keypoints with at least 50 STIPs within 15\,px, a proxy for static points on moving objects.
  Detection time is per camera, for the whole 300-frame clip on one RTX\,5090.
  Reprojection errors are measured over each point's inlier observations; matching and triangulation take 2--20\,s per scene.}
  \label{tab:recon}
\end{table*}

\begin{figure}[t]
  \centering
  \includegraphics[width=\linewidth]{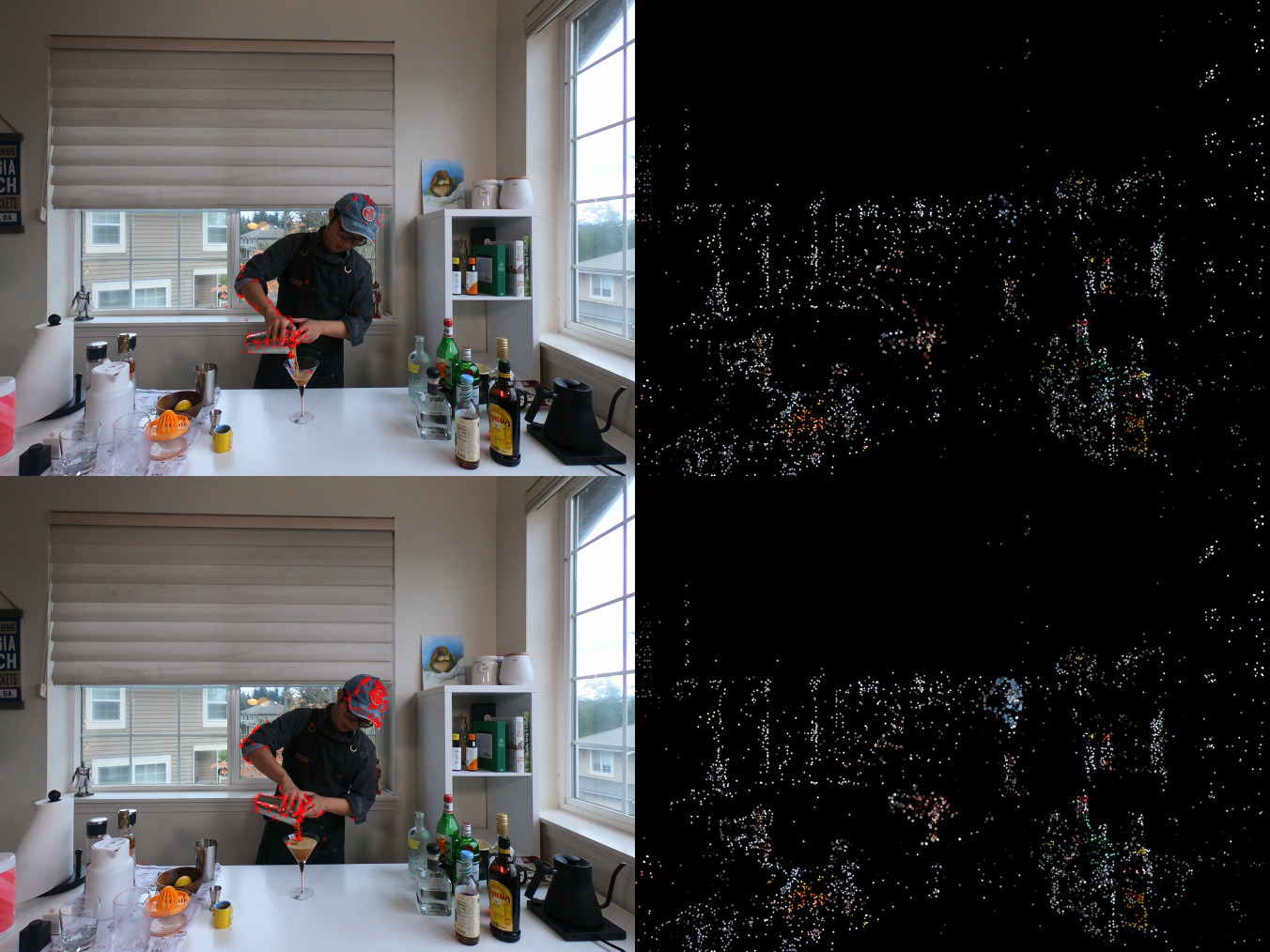}
  \caption{\textbf{\ours initialisation projected into held-out \texttt{cam00}} at frames 60 (top) and 200 (bottom). Left: dynamic points with $|t^\ast-f|\le0.5$ (red) over the ground-truth frame. Right: all points with their sampled colours. The camera was not used for reconstruction.}
  \label{fig:initcheck}
\end{figure}

\Cref{tab:recon} summarises \ours on all five scenes.
Detection is fast: about 5\,s per N3DV camera (1352$\times$1014$\times$300) with a 4\,GB peak of GPU memory, \ie 97--131\,s per scene.
Matching, tracking and triangulation then take 2--20\,s per scene.
The dynamic branch reconstructs 39k--113k time-stamped points per N3DV scene, each seen in about four views on average, with 0.65--0.83\,px mean reprojection error.
The static branch is far sparser, with about 3--5k points per scene, because large parts of the kitchens are low-texture blinds and walls.
\Cref{fig:initcheck} projects the \ours initialisation into the held-out camera: dynamic points with $|t^\ast - f|\le 0.5$ frames sit on the shaker, hands and cap at frame $f$.

\subsection{Robustness to matching thresholds}
\label{sec:sweep}

Without ground-truth geometry, we evaluate matching with a held-out camera.
We reconstruct without camera $h$ and project each point into $h$.
A point is \emph{verified} if one of $h$'s own detections lies within 3\,px of the projection, within $\pm1$ frame of $t^\ast$ for dynamic points, and with descriptor distance at most 1.0 to the point's mean descriptor.
\emph{Chance} is the same test with the projection shifted by 25\,px.
Precision, the fraction of visible points that are verified, is a lower bound: occlusion and missed detections also fail the test.

\begin{table}[t]
  \centering
  \small
  \setlength{\tabcolsep}{4pt}
  \begin{tabular}{@{}l cc r@{}}
    \toprule
    Scene / branch & Precision range & Chance & Best \# verified \\
    \midrule
    \texttt{coffee} static   & 0.31--0.37   & 0.04  & 1,377 \\
    \texttt{coffee} dynamic  & 0.424--0.450 & 0.03  & 40,128 \\
    \texttt{spinach} dynamic & 0.343--0.354 & 0.04  & 39,394 \\
    \bottomrule
  \end{tabular}
  \caption{\textbf{Held-out-camera verification} over the threshold grid. Precision varies by only $\sim$3 points across the whole grid and is 10--25$\times$ above chance.}
  \label{tab:sweep}
\end{table}

\Cref{tab:sweep} shows that matches are 10--25$\times$ above chance everywhere, and that the thresholds move precision by only about 3 points.
The epipolar, temporal and mutual gates do the filtering, so the ratio test is close to redundant.
The one setting that consistently matters is the dynamic epipolar gate. A 1\,px gate is best on every scene, with 2--3\% more verified points and 0.1--0.15\,px lower reprojection error than 3\,px.
This is consistent with sub-pixel calibration and sub-voxel STIP localisation.

\subsection{Initialising 4D Gaussian splatting}
\label{sec:4dgs}

\paragraph{Protocol.}
We use three N3DV scenes, \texttt{coffee\_martini}, \texttt{cook\_spinach} and \texttt{flame\_steak}, with the standard split: we train on all cameras except \texttt{cam00} and test on \texttt{cam00}.
Importantly, \emph{the initialisation never sees the test camera either}: \ours is re-run on the training cameras only.
This gives 4,860 static and 90,806 dynamic points on \texttt{coffee\_martini} (17 cameras, at 0.58 and 0.64\,px), 3,238 and 111,096 on \texttt{cook\_spinach}, and 3,027 and 84,595 on \texttt{flame\_steak} (20 cameras each).
We train 4DGS~\cite{yang2024real} with its N3DV configuration for 30k iterations, changing only the initialisation, and evaluate PSNR on all 300 test frames every 2k iterations.
We report the final (30k) model, not the best checkpoint, which would amount to selecting on the test set, and compute SSIM and LPIPS on it with the 4DGS evaluation code.
All implementations use same densification cap of 5.6M Gaussians, duw to limitation in GPU memory.
Training was done on a single NVIDIA RTX 5090 Blackwell GPU with 32 GB of VRAM.

\paragraph{Initialisations.}
\emph{Random}: 300k points sampled uniformly in the training frusta, with times uniform over the sequence (the 4DGS default).
\emph{COLMAP f0}: the standard first-frame cloud. We run COLMAP~\cite{schoenberger2016sfm} feature extraction, exhaustive matching and triangulation on frame 0 of the training cameras, with poses and intrinsics fixed to the training cameras (3,572 / 3,957 / 3,643 points; 0.75--0.77\,px; track length 4.7--5.4). Times are drawn as for random.
This is the sparse part of the official 4DGS N3DV preparation, without its MVS densification.
\emph{\ours}: our static and dynamic points (\cref{sec:init}); 88k--114k Gaussians.
\emph{\ours + COLMAP fill}: the COLMAP f0 points added to \ours as static Gaussians, initialised exactly like \ours's own static points.

\begin{table*}[t]
  \centering
  \small
  \setlength{\tabcolsep}{3.4pt}
  \resizebox{\linewidth}{!}{\begin{tabular}{@{}l ccccr ccccr ccccr@{}}
    \toprule
    & \multicolumn{5}{c}{\texttt{coffee\_martini}} & \multicolumn{5}{c}{\texttt{cook\_spinach}} & \multicolumn{5}{c}{\texttt{flame\_steak}} \\
    \cmidrule(lr){2-6}\cmidrule(lr){7-11}\cmidrule(lr){12-16}
    Initialisation & PSNR & 20--30k & SSIM & LPIPS & \#G & PSNR & 20--30k & SSIM & LPIPS & \#G & PSNR & 20--30k & SSIM & LPIPS & \#G \\
    \midrule
    Random              & 27.35 & 27.31 & .902 & .091 & 5.51M & 31.90 & 31.73 & .945 & .053 & 3.94M & 30.90 & 31.05 & .950 & .050 & 3.14M \\
    COLMAP f0           & 28.15 & 28.11 & .907 & .085 & 5.61M & 32.42 & 32.47 & .949 & .046 & 4.28M & 31.62 & 31.58 & .949 & .043 & 3.27M \\
    \ours               & 27.96 & 27.84 & .911 & .082 & \textbf{5.22M} & \textbf{32.55} & \textbf{32.56} & \textbf{.951} & .046 & \textbf{3.50M} & 31.00 & 30.86 & .952 & .044 & 2.85M \\
    \ours + COLMAP fill & \textbf{28.29} & \textbf{28.21} & \textbf{.912} & \textbf{.079} & 5.39M & 31.41 & 31.62 & .940 & \textbf{.045} & 3.57M & \textbf{32.02} & \textbf{32.23} & \textbf{.954} & \textbf{.040} & \textbf{2.73M} \\
    \bottomrule
  \end{tabular}}
  \caption{\textbf{4DGS on three N3DV scenes, held-out \texttt{cam00}} (all 300 frames). PSNR (dB), SSIM and LPIPS (AlexNet) of the final 30k model; ``20--30k'': PSNR averaged over the six evaluations from 20k to 30k; \#G: final number of Gaussians.}
  \label{tab:4dgs}
\end{table*}

\begin{figure*}[t]
  \centering
  \includegraphics[width=\linewidth]{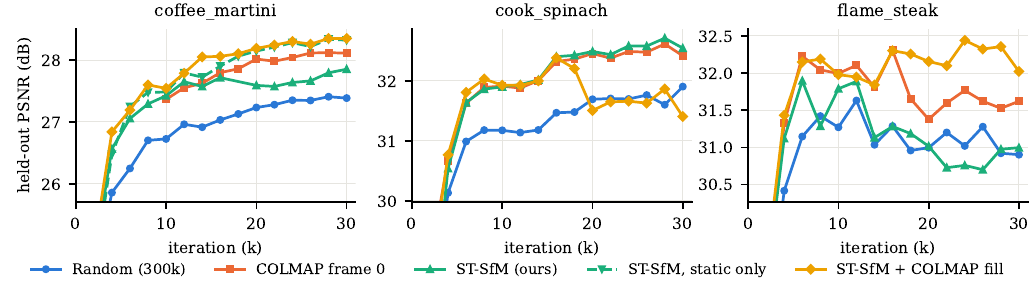}
  \caption{\textbf{Held-out PSNR during training} (\texttt{cam00}). \ours (green) is above random (blue) at every evaluation on \texttt{coffee\_martini} and \texttt{cook\_spinach}, and above or level with it on \texttt{flame\_steak}, where every arm peaks between 6k and 24k and then declines. The solid \ours \texttt{coffee\_martini} curve is the resumed run (evaluation at 18k lost); the dashed curve is the uninterrupted static-only variant.}
  \label{fig:curves}
\end{figure*}

\begin{figure*}[t]
  \centering
  \includegraphics[width=\linewidth]{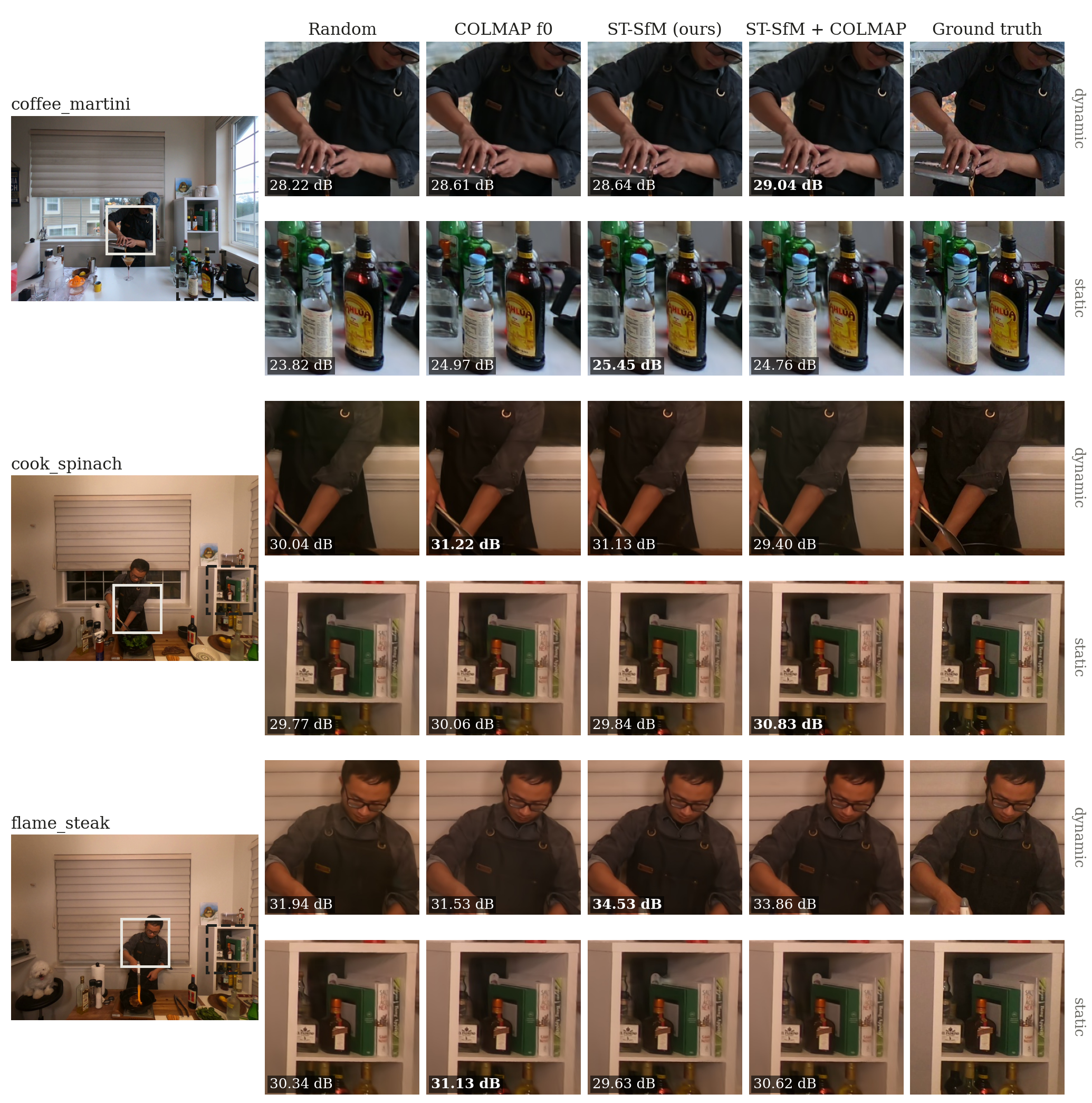}
  \caption{\textbf{Held-out \texttt{cam00} frame 150, final (30k) models.} Left: the ground-truth frame with the automatically chosen dynamic (solid) and static (dashed) boxes. Each crop is labelled with its own PSNR against ground truth; best in bold.}
  \label{fig:supp_crops}
\end{figure*}

\paragraph{Comparison with random initialisation.}
\Cref{tab:4dgs} and \cref{fig:curves} give the results.
\ours beats random initialisation on all three scenes and all three metrics. In PSNR the gain is +0.61, +0.64 and +0.09\,dB (on \texttt{coffee\_martini}: +0.47 and +0.76\,dB for the two seeds), starting from $2.6$--$3.4\times$ fewer Gaussians.
On \texttt{coffee\_martini} the lead is largest early: +0.70 and +0.99\,dB at 4k for the two seeds.
\ours also ends with the smallest model of the three single-source initialisations on every scene: 5\%, 11\% and 9\% fewer Gaussians than random, and 7\%, 18\% and 13\% fewer than COLMAP f0. \cref{fig:supp_crops} shows moving and static crops of all three scenes.

\paragraph{Comparison with first-frame COLMAP.}
Against the standard first-frame initialisation the picture is mixed and depends on the metric.
On \texttt{cook\_spinach}, \ours is ahead from 12k on and ends 0.13\,dB higher (+0.09\,dB over 20--30k), with slightly better SSIM, equal LPIPS and 18\% fewer Gaussians.
On \texttt{flame\_steak} it is 0.62\,dB lower in PSNR, but has higher SSIM (.952 \vs .949) and nearly equal LPIPS (.044 \vs .043). This scene is unstable for every arm: COLMAP f0 peaks at 32.30\,dB at 16k and random at 31.63\,dB at 12k before both decline, and \ours peaks at 31.89\,dB already at 6k.
On \texttt{coffee\_martini}, \ours leads through 8k (+0.54\,dB at 4k), COLMAP f0 is ahead from 10k, and at 30k \ours is 0.12\,dB lower in PSNR (28.07 \vs 28.19) but better in SSIM (.911 \vs .908) and LPIPS (.081 \vs .084).
In summary, \ours is ahead in SSIM on all three scenes, ahead in PSNR on one, and behind in PSNR on two as compared with COLMAP, by 0.12--0.19\,dB on \texttt{coffee\_martini} and 0.62\,dB on \texttt{flame\_steak}.

\begin{figure}[t]
  \centering
  \includegraphics[width=\linewidth]{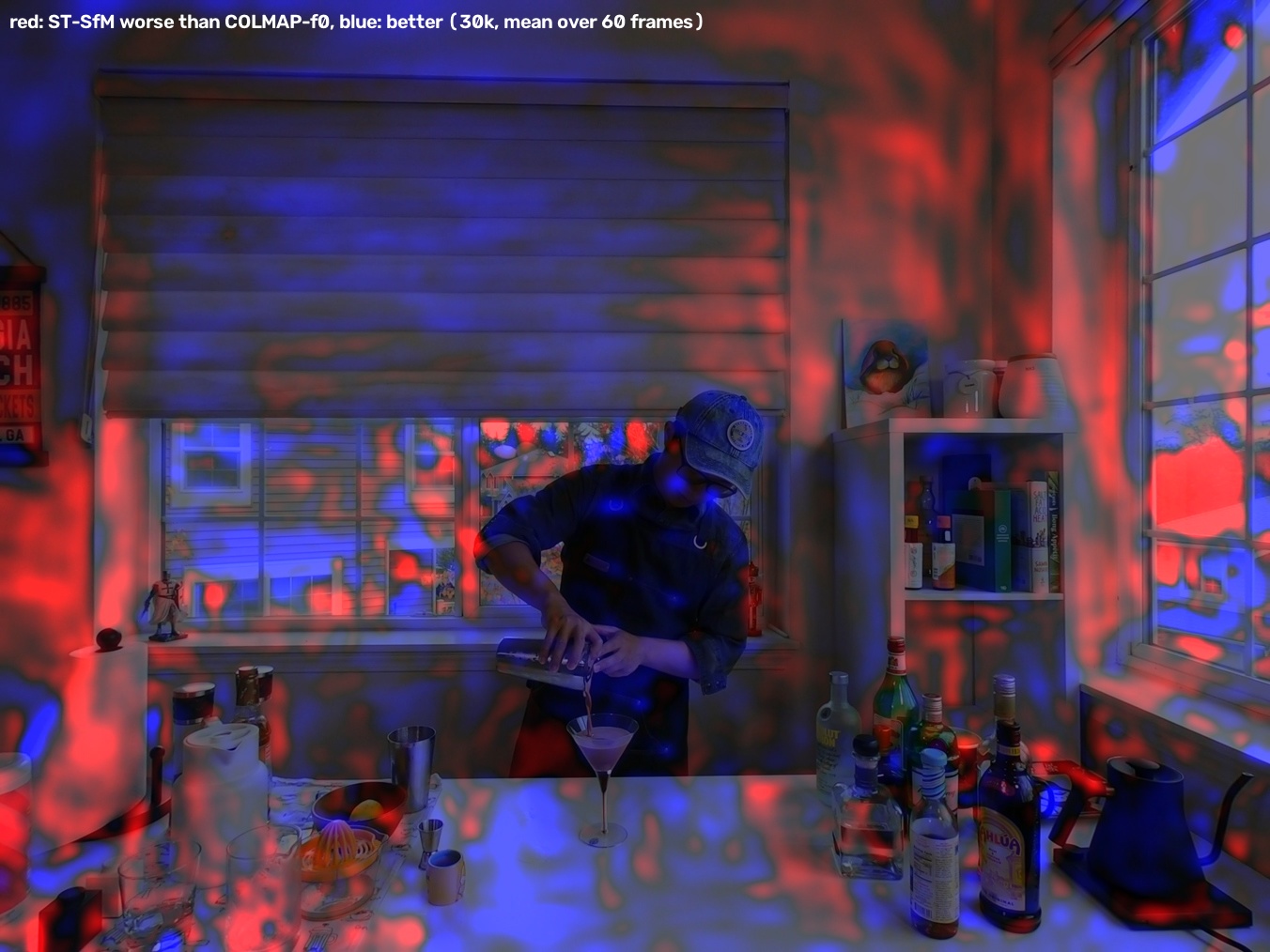}
  \caption{\textbf{Per-pixel error difference, \ours \vs COLMAP f0}, at 30k, averaged over 60 test frames. Red: \ours worse; blue: better.}
  \label{fig:errmap}
\end{figure}

\paragraph{Adding COLMAP points.}
Adding the COLMAP f0 points to \ours as static Gaussians is not consistently helpful.
It gives best results on \texttt{flame\_steak} (32.02\,dB, +0.40 over COLMAP f0, with the fewest Gaussians) and on \texttt{coffee\_martini} (28.35 and 28.22\,dB for the two seeds, +0.14\,dB over COLMAP f0 on average, with the best SSIM and LPIPS).
On \texttt{cook\_spinach} it is the worst at 30k in PSNR and SSIM (31.41\,dB), after dropping 0.9\,dB between 16k and 20k, although its LPIPS is the best.
In our single-scene analysis, we attributed the \ours--COLMAP gap on \texttt{coffee\_martini} to static coverage of the bright window exterior ( \cref{fig:errmap}).
\ours places more static points in that window region than COLMAP f0 does (326 \vs 208), and \ours reach the hybrid's PSNR without any COLMAP points.

\begin{table}[t]
  \centering
  \small
  \setlength{\tabcolsep}{3pt}
  \resizebox{\linewidth}{!}{\begin{tabular}{@{}l rr cc cc r@{}}
    \toprule
    & \multicolumn{2}{c}{Init points} & \multicolumn{2}{c}{PSNR (dB)} & & & \\
    \cmidrule(lr){2-3}\cmidrule(lr){4-5}
    Variant & static & dyn. & 10k & 30k & SSIM & LPIPS & \#G \\
    \midrule
    Default ($\delta{=}1$),   & 4,860 & 90,806 & 27.49 & 28.07 & .911 & .081 & \textbf{5.10M} \\
    $\delta = 0$                    & 4,860 & 4,298  & 27.56 & 28.33 & .912 & .082 & 5.31M \\
    $\delta = 2$                    & 4,860 & 91,077 & 27.46 & \textbf{28.39} & .912 & \textbf{.079} & 5.60M \\
    Dense static                    & 7,714 & 90,806 & \textbf{27.62} & 28.35 & .912 & \textbf{.079} & 5.22M \\
    Static only                     & 4,860 & 0      & 27.48 & 28.32 & \textbf{.913} & .081 & 5.27M \\
    Dynamic only                    & 0     & 90,806 & 26.85 & 27.34 & .892 & .102 & 5.68M \\
    Opacity reset off               & 4,860 & 90,806 & 22.43 & 21.44 & .765 & .287 & 5.66M \\
    \midrule
    Random                          & \multicolumn{2}{c}{300k} & 26.72 & 27.38 & .902 & .091 & 5.40M \\
    COLMAP f0                       & 3,572 & --     & 27.36 & 28.11 & .906 & .086 & 5.60M \\
    \ours + COLMAP fill             & 8,432 & 90,806 & 27.54 & 28.35 & .912 & .080 & 5.61M \\
    \bottomrule
  \end{tabular}}
  \caption{\textbf{Ablations on \texttt{coffee\_martini}} (held-out \texttt{cam00}). Top: \ours variants, best in bold. Bottom: the other initialisation schemes of \cref{tab:4dgs}.}
  \label{tab:ablation}
\end{table}

\subsection{Ablations}
\label{sec:ablations}

\Cref{tab:ablation} varies one component at a time on \texttt{coffee\_martini}  from N3DV dataset.

\paragraph{Temporal window $\delta$.}
With $\delta = 0$, only STIPs with identical sub-frame times can match, and the dynamic branch keeps only 4,298 points, 21$\times$ fewer than with $\delta = 1$. With $\delta = 2$ it gains under 1\% over $\delta = 1$.
At 30k, $\delta = 0$ and $\delta = 2$ differ by only 0.06\,dB.
On this scene, therefore, the final PSNR is not sensitive to the \emph{number} of dynamic points over this range. Our approach builds upon and adapts an existing 4D Gaussian Splatting (4DGS) optimisation framework. Which suggests that its underlying optimisation strategy is not designed to fully leverage the information encoded in our input representation when optimising the 4D Gaussians.

\paragraph{Static and dynamic branches.}
The static branch alone, 4,860 points, reaches 28.32\,dB at 30k with the best SSIM of all variants.
The dynamic branch alone, 90,806 points, reaches 27.34\,dB, level with random initialisation from $3.3\times$ fewer points. With no background initialised it starts far behind (23.56 \vs 25.86\,dB at 4k), and it overtakes random by 10k.
Adding the dynamic points to the static ones changes PSNR at 10k by $-0.06$ to $+0.14$\,dB and LPIPS at 30k by $-0.002$ to $+0.002$, which is within run-to-run variation: with the same seed and initialisation.

\paragraph{Opacity reset.}
Our initial plan was to disable the periodic opacity reset of 4DGS for sparse initialisations.
This turns out to be harmful.
Without it, held-out PSNR follows the default run up to 4k (26.68\,dB), then collapses to 22.43\,dB by 10k and 21.44\,dB at 30k (SSIM .765, LPIPS .287), with 5.66M Gaussians.
In this 4DGS implementation, the reset interval also gates pruning of large screen-space Gaussians, so disabling it removes both. We keep the default reset for all implementations.

\section{Limitations and Conclusion}
\label{sec:conclusion}

\paragraph{Limitations.}
\ours assumes synchronised and calibrated input cameras, which limits its applicability to settings where these requirements are not satisfied. Currently, we reconstruct 4D sparse clouds, which are subsequently used for initialisation. Our current approach builds upon and adapts an existing 4DGS optimisation framework. However, this dependency introduces certain limitations, as the existing optimisation scheme cannot fully exploit the information available in our input representation during the optimisation of 4D Gaussians. As future work, we aim to develop a dedicated optimisation pipeline tailored to our input structure, drawing inspiration from \ours, to better leverage the available information and achieve more effective 4D Gaussian optimisation.

\paragraph{Conclusion.}
\ours treats space-time interest points as correspondences for synchronised multi-camera video. Matched epipolarly and in time, chained into consistent tracks and triangulated robustly, they give a sparse, sub-pixel-accurate, time-stamped 4D reconstruction of what moves, in seconds per scene.
As an initialisation for 4D Gaussian splatting, it consistently improves on random points, from far fewer initial Gaussians, and is competitive with the standard first-frame SfM cloud with a smaller final model.
{
    \small
    \bibliographystyle{ieeenat_fullname}
    \bibliography{main}
}

\end{document}